\documentclass{article}
\usepackage{spconf,amsmath,graphicx,hyperref}
\usepackage{multirow}
\usepackage{pifont}   
\usepackage{hyperref}

\title{FusionEdit: Semantic Fusion and Attention Modulation for Training-Free Image Editing}
\name{Yongwen Lai$^{1}$,
      Chaoqun Wang$^{1\dagger}$\thanks{$\dagger$ Corresponding author.},
      Shaobo Min$^{2}$}
\address{
    $^{1}$School of Artificial Intelligence, South China Normal University, Guangzhou, China\\
    $^{2}$University of Science and Technology of China, Hefei, China
}

\begin{document}
\ninept
\maketitle
\begin{abstract}
Text-guided image editing aims to modify specific regions according to the target prompt while preserving the identity of the source image. 
Recent methods exploit explicit binary masks to constrain editing, but hard mask boundaries introduce artifacts and reduce editability. 
To address these issues, we propose FusionEdit, a training-free image editing framework that achieves precise and controllable edits. 
First, editing and preserved regions are automatically identified by measuring semantic discrepancies between source and target prompts. 
To mitigate boundary artifacts, FusionEdit performs distance-aware latent fusion along region boundaries to yield the soft and accurate mask, and employs a total variation loss to enforce smooth transitions, obtaining natural editing results. 
Second, FusionEdit leverages AdaIN-based modulation within DiT attention layers to perform a statistical attention fusion in the editing region, enhancing editability while preserving global consistency with the source image. 
Extensive experiments demonstrate that our FusionEdit significantly outperforms state-of-the-art methods. Code is available at \href{https://github.com/Yvan1001/FusionEdit}{https://github.com/Yvan1001/FusionEdit}.
\end{abstract}

\section{Introduction}
\label{sec:intro}

Text-guided image editing~\cite{huang2024entwinedinversion, zhan2025madiff, lau2025genie, huang2025diffusion} has gained increasing attention in computer vision, enabling flexible modification of images according to textual instructions while preserving the identity and structure of the source image. 
With the progress of generative models~\cite{rombach2022high, saharia2022photorealistic,peebles2023scalable}, particularly recent rectified flow models such as FLUX~\cite{flux2024} and SD3~\cite{esser2024scaling}, high-quality image synthesis is now feasible. 
However, adapting these powerful models for controllable and precise editing remains significantly challenging.

Existing image editing approaches \cite{hertz2023prompttoprompt,zhu2025kv,wang2024taming,rout2024semantic} typically balance two objectives, \emph{i.e.}, preserving source content and aligning with the target prompt. 
For source preservation, P2P~\cite{hertz2023prompttoprompt} injects cross-attention maps from reconstruction branch into editing branch. RF-Solver-Edit~\cite{wang2024taming} and PnP~\cite{tumanyan2023plug} reuses self-attention layer values from source inversion. 
For target alignment, RF Inversion~\cite{rout2024semantic} employs a time-dependent controller to adjust target semantic strength during denoising. FluxSpace~\cite{dalva2024fluxspace} enables disentangled target semantic editing via leveraging their learned representations in rectified flow models.
Despite these advances, these methods lack spatial constraints, introducing unintended modifications outside the desired regions. 
To address this issue, recent works incorporate explicit binary masks from blend words or segmentation models to localize editing regions~\cite{cao2023masactrl,  soni2025locatedit, hu2025dcedit, shagidanov2024groundedinstruct, swami2025promptartisan}. 
For example, LOCATEdit~\cite{soni2025locatedit} explicitly uses blend words as additional input and employs graph-based approach to enhance cross-attention maps. 
DCEdit~\cite{hu2025dcedit} leverages visual and textual self-attention to enhance the cross-attention maps for editing region localization.
While binary masks provide spatial guidance, their rigid region boundaries inevitably produce hard artifacts and limit editability due to the strict separation of editing and preserved regions.

In this paper, we propose FusionEdit, a novel training-free framework to achieve precise and controllable image editing. 
FusionEdit first automatically determines editing regions by quantifying semantic discrepancies between source and target prompts, eliminating the reliance on external masks. 
To suppress boundary artifacts, FusionEdit introduces distance-aware latent fusion along region transitions, combined with a total variation (TV) loss to ensure smooth and coherent soft mask. 
Furthermore, FusionEdit designs disparity-aware attention modulation (DAM), integrating AdaIN-based~\cite{huang2017arbitrary} modulation into DiT attention layers, enabling statistical fusion of source appearance and target semantics to enhance editability while preserving global consistency. 
Extensive experiments on the challenging benchmark demonstrate that FusionEdit achieves superior results compared with state-of-the-art methods. The main contributions of this work are as follows:
\begin{itemize}
\item We propose FusionEdit, a training-free text-guided image editing framework that achieves precise and controllable editing.
\item We design semantic-discrepancy-based localization, using distance-aware latent fusion with soft mask for smooth and natural editing, and introduce AdaIN-based attention modulation to enhance editability and global semantic consistency.
\item Comprehensive experiments show that FusionEdit establishes state-of-the-art performance on the challenging benchmark.
\end{itemize}

\begin{figure*}[h]
    \centering
    \includegraphics[width=\textwidth]{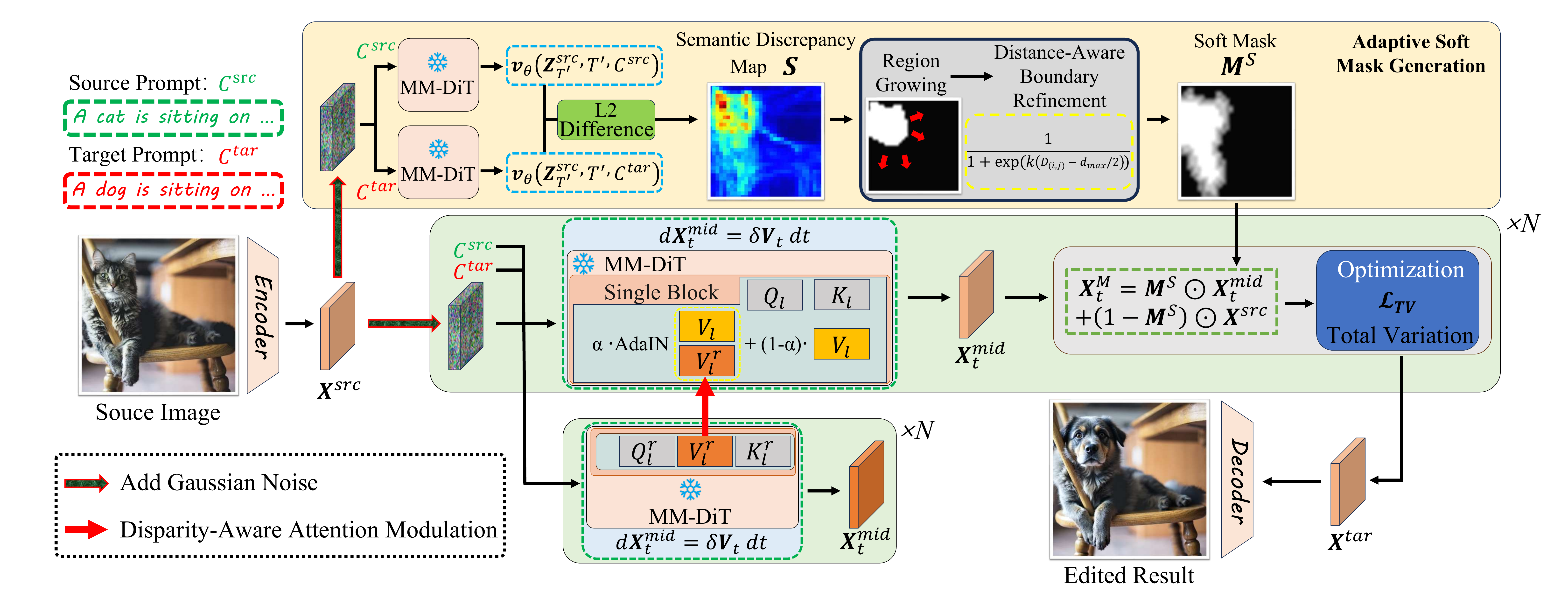}
    \caption{Overview of the FusionEdit pipeline. Given a source image with source and target prompts, FusionEdit generates the semantic discrepancy map from target- and source-conditioned velocity fields to produce an adaptive soft mask that guides localized editing. Disparity-aware attention modulation (DAM) further injects global appearance statistics from the unmasked stream into the masked editing path, enabling precise and consistent image editing.}
    \label{fig:pipeline}
    \vspace{-0.5cm}
\end{figure*}

\section{Preliminaries}
\label{sec:prelim}

Rectified flow (RF) models~\cite{liu2022flow} establish a continuous mapping between the image data distribution $\boldsymbol{X}$ and a Gaussian prior $\boldsymbol{N} \sim \mathcal{N}(0, I)$ via an ordinary differential equation (ODE):
\begin{equation}
d\boldsymbol{X}_t = \boldsymbol{v}_{\theta}(\boldsymbol{X}_t, t, C)\, dt,
\label{eq:rf_ode}
\end{equation}
where $\boldsymbol{v}_{\theta}$ is a pretrained velocity field, $t \in [0,1]$ denotes the timestep variable, and $C$ is the conditioning text prompt. A key property of RF is that the marginal distribution satisfies a linear interpolation $\boldsymbol{X}_t \sim (1-t)\boldsymbol{X} + t\boldsymbol{N}$, which enables efficient sampling.

In text-guided image editing, given a source latent $\boldsymbol{X}^{src}$ with prompt $C^{src}$ and a target prompt $C^{tar}$, the goal is to generate an edited latent $\boldsymbol{X}^{tar}$ that aligns with $C^{tar}$ while preserving the identity of $\boldsymbol{X}^{src}$. 
Recent methods~\cite{kulikov2024flowedit, kim2025flowalign,xie2025dnaedit} leverage pretrained RF models to construct an ODE trajectory from the source to the target distribution. 
Let $\boldsymbol{X}_t^{mid}$ denote the intermediate latent along this trajectory, with initial condition $\boldsymbol{X}_1^{mid} = \boldsymbol{X}^{src}$ and final result $\boldsymbol{X}_0^{mid} = \boldsymbol{X}^{tar}$. Its evolution is given by:
\begin{equation}
d\boldsymbol{X}_t^{mid} = \delta\boldsymbol{V}_t \, dt,
\label{eq:delta_v}
\end{equation}
where $\delta \boldsymbol{V}_t = 
\boldsymbol{v}_{\theta}(\boldsymbol{Z}_t^{tar}, t, C^{tar})
- \boldsymbol{v}_{\theta}(\boldsymbol{Z}_t^{src}, t, C^{src})$ represents the semantic displacement field between the target and source prompts. Here:
\begin{equation}
\boldsymbol{Z}_t^{src} = (1-t)\boldsymbol{X}^{src} + t\boldsymbol{N}, 
\quad
\boldsymbol{Z}_t^{tar} = \boldsymbol{X}_t^{mid} - t\boldsymbol{X}^{src} + t\boldsymbol{N},
\end{equation}
where $\boldsymbol{Z}_t^{src}$ denotes the forward-noised source latent.
$\boldsymbol{Z}_t^{tar}$ projects the intermediate latent $\boldsymbol{X}_t^{mid}$ into the same noise space, which allows direct comparison of source- and target-conditioned velocities.
By solving the ODE backward from $t=1$ to $t=0$, the final edited result $\boldsymbol{X}^{tar}$ is obtained.

However, this process does not specify where the edits should occur. 
To enforce spatial control, mask-based methods \cite{soni2025locatedit, hu2025dcedit} introduce the binary mask $\boldsymbol{M}$ to restrict editing to specific regions:
\begin{equation}
\tilde{\boldsymbol{X}}^{tar} = \boldsymbol{M} \odot \boldsymbol{X}^{tar} + (1-\boldsymbol{M}) \odot \boldsymbol{X}^{src},
\end{equation}
where $\odot$ denotes element-wise multiplication. 
Although effective, binary masks produce rigid boundaries that lead to visual artifacts and reduced editability due to the strict separation of editing and preserved regions.

\section{Method}
In this paper, we propose a training-free framework termed FusionEdit for text-guided image editing. 
It consists of two key components: 1) adaptive soft mask generation for distance-aware latent fusion, which accurately localizes editing regions with smooth boundaries, and 2) disparity-aware attention modulation (DAM) with AdaIN, which injects global source statistics to preserve visual consistency. 
The overall pipeline is illustrated in Fig.~\ref{fig:pipeline}.

\subsection{Adaptive Soft Mask Generation}

Accurately localizing editing regions is crucial for controllable text-guided editing. 
Existing approaches often rely on external knowledge, \emph{e.g.}, blend words or segmentation models, and produce rigid binary masks, inevitably causing boundary artifacts and restricting editability. 
To this end, we propose an adaptive soft mask generation scheme that automatically identifies semantic editing regions without explicit knowledge and enforces smooth transitions along region boundaries.

\noindent
\textbf{Region Extraction via Semantic Discrepancy.}  
We observe that the difference between source and target prompts naturally indicates regions to edit. 
To capture this, we compute a semantic discrepancy map by contrasting target- and source-conditioned velocity fields at an intermediate timestep $T'$, which directly calculates the pixel-wise L2 distance between these two velocity fields:
\begin{equation}
\boldsymbol{S} =\| \boldsymbol{v}_{\theta}
(\boldsymbol{Z}^{src}_{T'},T', C^{tar}) - \boldsymbol{v}_{\theta}(\boldsymbol{Z}^{src}_{T'},T', C^{src})\|_2.
\end{equation}
Here, $T'$ is chosen in the early-to-middle denoising stage, where semantic information is still preserved and not overwhelmed by noise, making the discrepancy reliable.
To reduce randomness, we repeat the process several times and average the results, obtaining a stable map $\bar{\boldsymbol{S}}$ that indicates potential editing regions.

Instead of pixel-wise thresholding, which is noisy, we divide $\bar{\boldsymbol{S}}$ into non-overlapping patches and compute the mean discrepancy per patch. 
Starting from the patch with maximum discrepancy, neighboring patches with similar mean values are iteratively merged. 
This region-growing strategy ensures spatially coherent editing regions, forming a binary mask $\boldsymbol{M}^R$.
Compared with $k$-means or Otsu thresholding, our method avoids fragmented or noisy regions and yields the mask aligned with semantic boundaries.

\noindent
\textbf{Soft Boundary Refinement.}  
Although $\boldsymbol{M}^R$ localizes editing regions, its binary nature causes abrupt transitions.
To mitigate this, we convert $\boldsymbol{M}^R$ into a soft mask $\boldsymbol{M}^S$ via distance-aware strategy and employ latent fusion combined with TV loss. Unlike binary masks, this soft mask can achieve a smooth transition band in the boundary area, which enables a smoother and more natural fusion of the latents.
 
Formally, for each pixel $(i,j)$:
\begin{equation}
\boldsymbol{M}^S_{(i,j)} = 
\begin{cases} 
\frac{1}{1 + \exp \big(k \,(D_{(i,j)} - d_{\text{max}}/2)\big)}, & \text{if } D_{(i,j)}\leq d_{\text{max}},\\
\quad \quad \quad \boldsymbol{M}^R_{(i,j)} \quad\quad\quad\quad
, & \text{if } D_{(i,j)}>d_{\text{max}},
\end{cases}
\end{equation}
where $D_{(i,j)}$ is the distance of pixel $(i,j)$ to the nearest boundary, $d_{\text{max}}$ controls the width of the transition band, and $k$ adjusts boundary sharpness. Pixels within $D_{(i,j)} \leq d_{\text{max}}$ form a smooth transition band, while others remain identical to the binary mask.
The fused latent at timestep $t$ is then expressed as:
\begin{equation}
\boldsymbol{X}^M_t = \boldsymbol{M}^S \odot \boldsymbol{X}_t^{mid} + (1 - \boldsymbol{M}^S) \odot \boldsymbol{X}^{src}.
\label{eq:weighted_fusion}
\end{equation}
This design ensures a smooth transition between the editing and preserved regions.
To further eliminate artifacts, we optimize the fusion by minimizing a total variation loss:
\begin{equation}
\mathcal{L}_{TV} = 
\sum_{(i,j) \in \Omega_b} \|\nabla \boldsymbol{X}^M_{t\,(i,j)}\|^2 
+ \lambda \sum_{(i,j) \in \Omega_b} \|\boldsymbol{X}^M_{t\,(i,j)} - \hat{\boldsymbol{X}}^M_{t\,(i,j)}\|^2,
\end{equation}
where $\Omega_b$ denotes the boundary region, $\nabla \boldsymbol{X}^M_t$ enforces spatial smoothness, $\hat{\boldsymbol{X}}^M_t$ denotes the initial fusion result from Eq.~\ref{eq:weighted_fusion} at timestep $t$, and $\lambda$ balances smoothness with fidelity.

Through semantic discrepancy-based localization and soft boundary refinement, the proposed mask retains the fidelity of preserved regions while enabling smooth and natural transitions in editing regions.

\begin{figure}[t]
	\centering
	\includegraphics[width=1\linewidth]{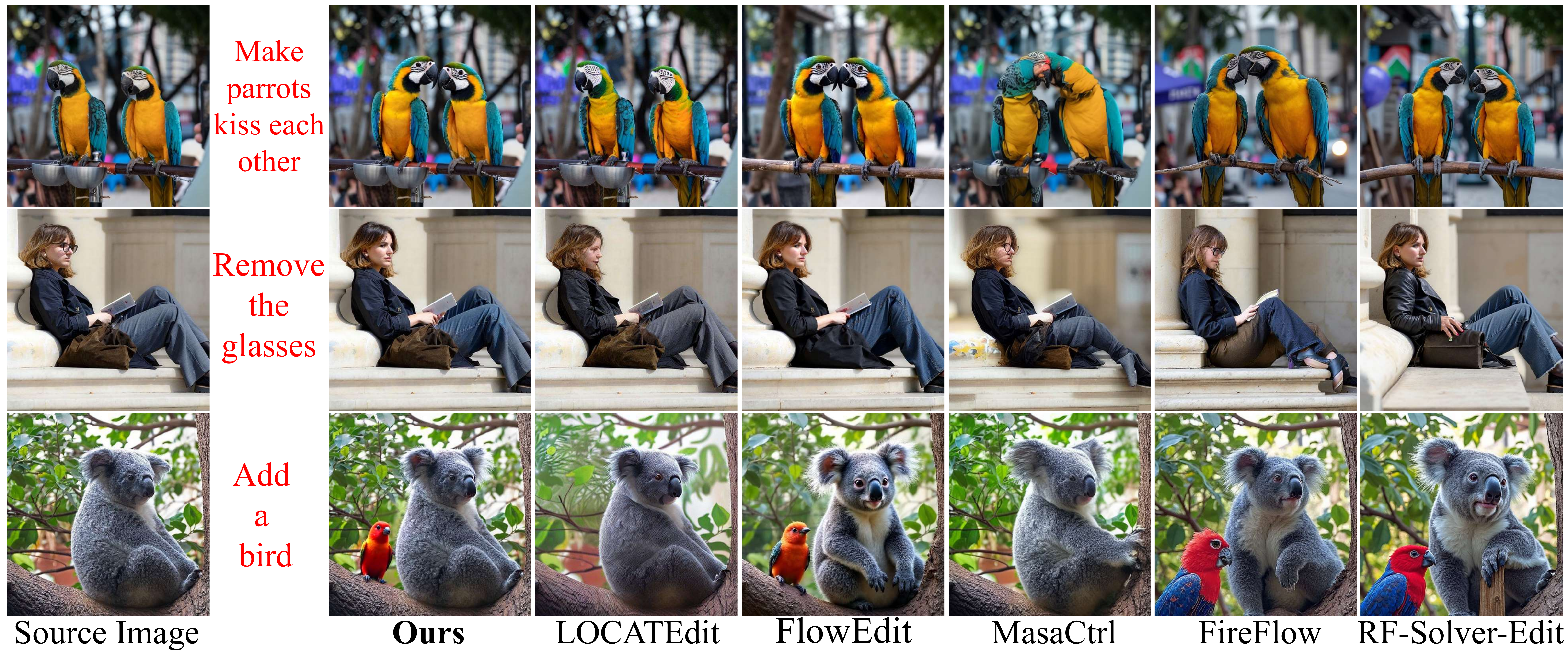}
	\caption{Qualitative comparisons on text-guided image editing. Our FusionEdit achieves faithful modifications according to the target prompt while preserving essential content of the source image.}
	\label{fig:sota}
\end{figure}

\subsection{Disparity-Aware Attention Modulation}

While the soft mask confines editing to intended regions, its spatial separation still suppresses cross-region semantic cues and ignores the global style information that guides fine-grained alignment.  
To this end, we introduce a disparity-aware attention modulation (DAM) module that adaptively injects global appearance statistics captured from an unmasked generation stream into the masked target editing process.
These global statistics provide a compact description of the source-level visual semantics.
By supplementing the masked editing path with these statistics, DAM restores contextual cues that masking suppresses, thereby alleviating the loss of fine-grained editability while keeping the overall visual appearance consistent with source.

Let $V_l$ denote the value tensor of the masked editing path and $V_l^{r}$ the value tensor obtained from a parallel reference path that follows the same denoising process but without mask constraints at $l$-th layer (timestep $t$ omitted for clarity).  
We fuse them through adaptive instance normalization (AdaIN)~\cite{huang2017arbitrary}:
\begin{equation}
\label{eq:fuseadain}
V_l' = \alpha \cdot \text{AdaIN}(V_l, V_l^{r}) + (1-\alpha) \cdot V_l,
\end{equation}
where:
\begin{equation}
\text{AdaIN}(V_l, V_l^{r}) =
\sigma(V_l^{r}) \left( \frac{V_l - \mu(V_l)}{\sigma(V_l)} \right) + \mu(V_l^{r}),
\end{equation}
and $\mu(\cdot)$, $\sigma(\cdot)$ are the channel-wise mean and standard deviation.  
This operation transfers the global feature statistics of the unmasked stream into the masked editing branch, enriching the semantic context while preserving the spatial attention pattern dictated by the mask.

To control the strength of this fusion, we leverage the adaptive weight $\alpha$:
\begin{equation}
\alpha = \beta \cdot (1 - t)\cdot \bigl[ 1 - \gamma \cdot(\bar{\Delta} - \eta) \bigr],
\end{equation}
where $\bar{\Delta}$ is the mean magnitude of $\bar{\boldsymbol{S}}$, which quantifies the overall discrepancy between source and target prompts.
$\beta$ is the base strength.
$\gamma,\eta$ control sensitivity and neutral threshold.
We clip $\alpha$ to $[0,1]$ for valid interpolation.  
This adaptive design ensures that DAM injects global statistics strengthening detail recovery when prompts are close and restraining fusion when large semantic changes are required.

\begin{table}[t]
 \caption{Quantitative results on text-guided image editing. The best result is bolded, and the second best is underlined. ``*'' denotes mask-based methods.}
    \resizebox{\linewidth}{!}{  
    \begin{tabular}{c||c||c|c|c|c||c}
        \hline
        \multirow{2}{*}{Method}
        &\multicolumn{1}{c||}{Structure} 
        & \multicolumn{4}{c||}{Background Preservation} 
        & \multicolumn{1}{c}{CLIP Similariy}  \\ \cline{2-7}
         
        &$\text{Distance}_{\times10^3}$ $\downarrow$
        & PSNR $\uparrow$
        &$\text{LPIPS}_{\times10^2}$ $\downarrow$
        & $\text{MSE}_{\times10^2}$ $\downarrow$
        &$\text{SSIM}_{\times10^2}$ $\uparrow$
        & Whole $\uparrow$\\ 
         
         \hline \hline

        RF Inversion \cite{rout2024semantic} 
        & 41.69 
        & 20.6885 
        & 18.56  
        & 1.26
        & 70.78 
        & \underline{25.5823}
        \\ \hline

        InfEdit
        \cite{xu2024inversion}
        & 32.04 
        & 21.7840 
        & 11.06  
        & 1.95
        & 77.84
        & 25.5277 
        \\ \hline

        FlowEdit\cite{kulikov2024flowedit}  
        & 22.80 
        & 23.8565
        & \underline{8.12}  
        & 0.63
        & \underline{86.94}
        & 25.5106
        \\ \hline

        LOCATEdit\cite{soni2025locatedit}*   
        & 38.46 
        & 21.8626
        & 11.94
        & 1.13
        & 80.48
        & 24.8166
        \\ \hline
        
        MasaCtrl\cite{cao2023masactrl}*  
         & 27.53
         & 22.3053
         & 10.25
         & 0.84
         & 80.41
         & 24.9633
        
        \\ \hline

         RF-Solver-Edit\cite{wang2024taming} + DCEdit \cite{hu2025dcedit}*
         & 26.98 
         & 24.4400 
         & 11.40
         & 0.59
         & 83.45
         & 25.3400
         \\ \hline

        FireFlow\cite{deng2024fireflow} + DCEdit\cite{hu2025dcedit}*
        & \underline{22.36} 
        & \underline{25.4100}
        & 9.40  
        & \textbf{0.48} 
        & 85.00 
        & 25.4700
        \\ \hline  \hline

        \textbf{FusionEdit (Ours)}*  
        & \textbf{14.52} 
        & \textbf{25.6451} 
        & \textbf{5.14} 
        & \underline{0.53} 
        & \textbf{90.91} 
        & \textbf{25.6491}  
        \\ \hline
    \end{tabular}
    } 
   
	\label{tab:sota_result}
\end{table}

\begin{figure}[t]
    \centering
    \includegraphics[width=\linewidth]{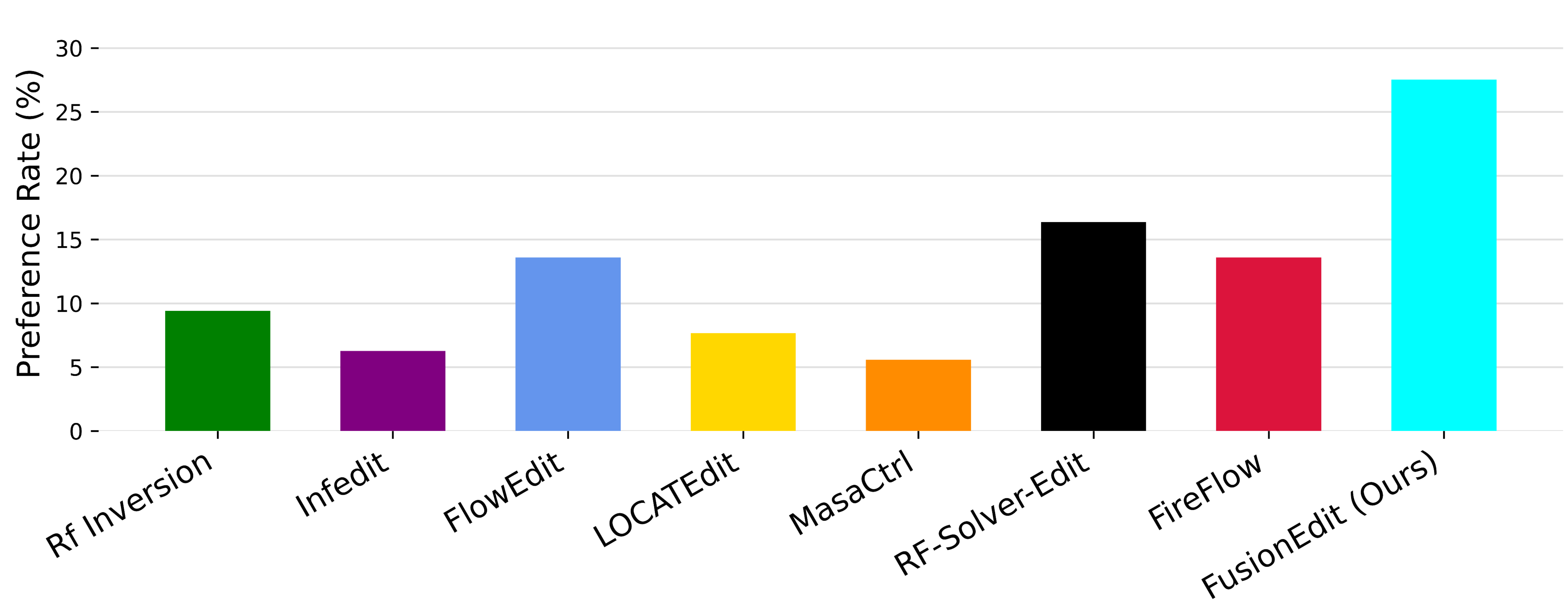}
    \caption{User study results. Our FusionEdit receives the highest number of user selections.}
    \label{fig:userstudy}
     \vspace{-0.5cm}
\end{figure}

\section{Experimental Results}
\subsection{Experimental Settings}
\textbf{Dataset Details.} 
We evaluate our method on PIE-Bench~\cite{ju2024pnp}, which is a comprehensive benchmark containing $700$ images across diverse scenes and editing categories.  
Each editing instance consists of a source image with its corresponding source prompt and a target prompt, enabling systematic assessment of editing fidelity and controllability.

\noindent
\textbf{Implementation Details.} We adopt Flux~\cite{flux2024} as rectified flow backbone.
Following~\cite{kulikov2024flowedit}, we set the total timestep to $28$, with source and target guidance values of $1.5$ and $5.5$, respectively.
The hyperparameters are set as follows: intermediate timestep $T'=0.89$; transition band width \(d_{\text{max}}=3\); boundary sharpness $k=5$; smoothness-fidelity trade-off $\lambda=50$; base fusion strength $\beta=0.1$; disparity sensitivity $\gamma=0.5$; and neutral threshold $\eta=0.5$. All experiments are conducted on an NVIDIA 5880 GPU with 48 GB memory.

\noindent
\textbf{Evaluation Metrics.} Following previous studies~\cite{soni2025locatedit, hu2025dcedit,ju2024pnp}, we employ CLIP-T~\cite{radford2021learning}, PSNR, LPIPS~\cite{zhang2018unreasonable}, MSE, SSIM~\cite{wang2004image}, and structure distance~\cite{tumanyan2022splicing} for evaluation. Specifically, CLIP-T assesses alignment between the edited image and the target prompt, while other metrics evaluate content preservation of preserved regions by comparing the edited image with the source image.
 
\subsection{Comparison with State-of-the-art Methods}
\noindent \textbf{Text-Guided Image Editing.} We compare our FusionEdit with recent state-of-the-art approaches for text-guided image editing. Quantitative results, shown in Table~\ref{tab:sota_result}, indicate that our method achieves excellent performance across most metrics. 
This highlights the effectiveness of our soft fusion strategy and disparity-aware attention modulation.
Notably, while DCEdit~\cite{hu2025dcedit} explicitly utilizes additional blend words as input, FusionEdit surpasses it without requiring extra supervision.
Qualitative results in Fig.~\ref{fig:sota} further confirm these findings: FusionEdit achieves higher semantic accuracy in editing regions while better preserving background fidelity, leading to more natural and coherent results.

\noindent \textbf{User Study.} 
To complement quantitative evaluation, we conduct a user study assessing the perceptual quality of FusionEdit against competing methods.  
A total of $50$ participants were presented with the source image, source prompt, and target prompt, together with editing results from different methods displayed in randomized order.  
For each case, participants were asked to select the most visually satisfactory result, with a ``Not Sure'' option provided to avoid forced choices.  
Each participant evaluated approximately $10$ sets.  
As shown in Fig.~\ref{fig:userstudy}, FusionEdit receives the highest preference rate.

\begin{table}
	\centering
    \caption{Quantitative effect of adaptive soft mask on editing performance.}
\centering
    \resizebox{\linewidth}{!}{  
    \begin{tabular}{c||c||c|c|c|c||c}
        \hline
        \multirow{2}{*}{Method}
        &\multicolumn{1}{c||}{Structure} & \multicolumn{4}{c||}{Background Preservation} & \multicolumn{1}{c}{CLIP Similariy}  \\ \cline{2-7}
        &$\text{Distance}_{\times10^3}$ $\downarrow$&PSNR $\uparrow$&$\text{LPIPS}_{\times10^2}$ $\downarrow$& $\text{MSE}_{\times10^2}$ $\downarrow$&$\text{SSIM}_{\times10^2}$ $\uparrow$& Whole $\uparrow$\\ \hline \hline

        W/o Soft Mask 
        & 22.80 
        & 23.8565
        & 8.12
        & 0.63
        & 86.94
        & 25.5106
        \\ \hline

        W/ Soft Mask 
        & \textbf{15.02} 
        & \textbf{25.9888}
        & \textbf{4.95}
        & \textbf{0.46}
        & \textbf{91.15}
        & \textbf{25.5197}
        \\ \hline

    \end{tabular}
    } 
	\label{tab:ablation1}
\end{table}

\begin{figure}[t]
    \centering
    \includegraphics[width=\linewidth]{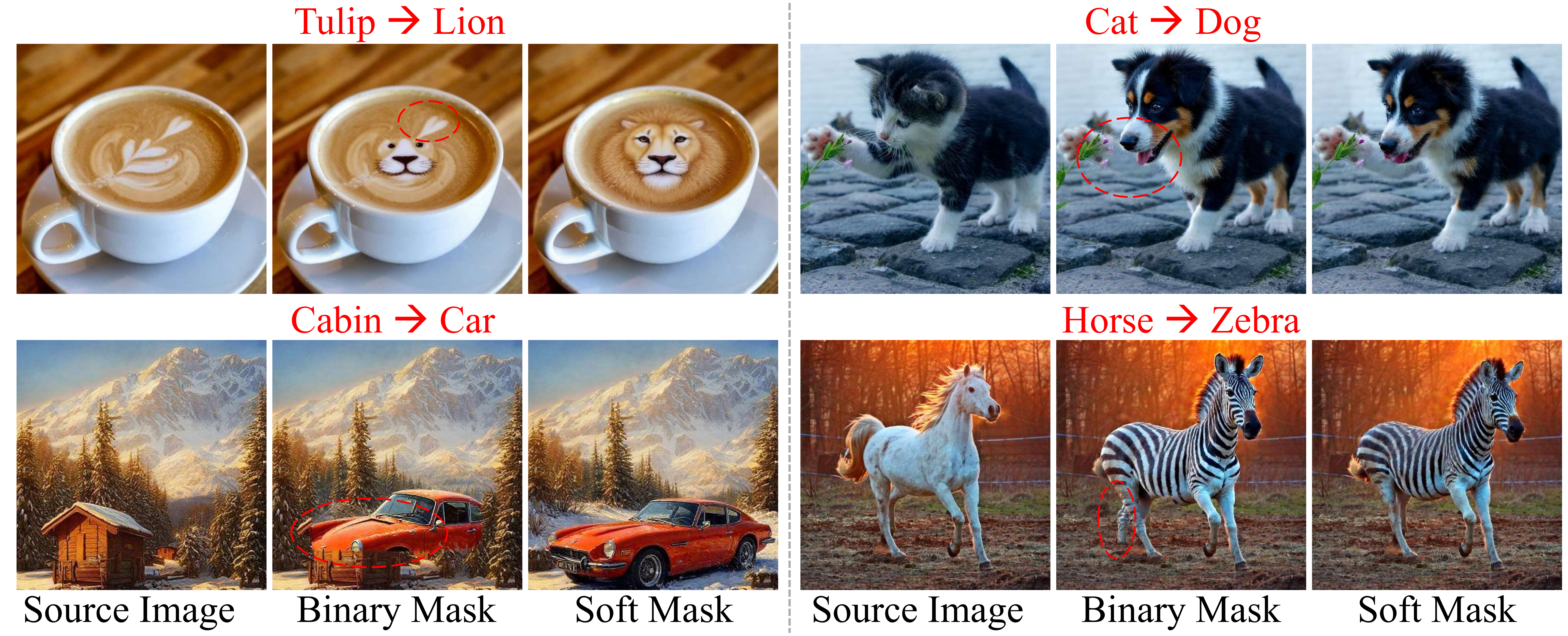}
    \caption{Comparison of editing results with binary mask vs. soft mask.}
    \vspace{-0.4cm}
    \label{fig:ablation2}
\end{figure}

\begin{figure}[t]
    \centering
    \includegraphics[width=\linewidth]{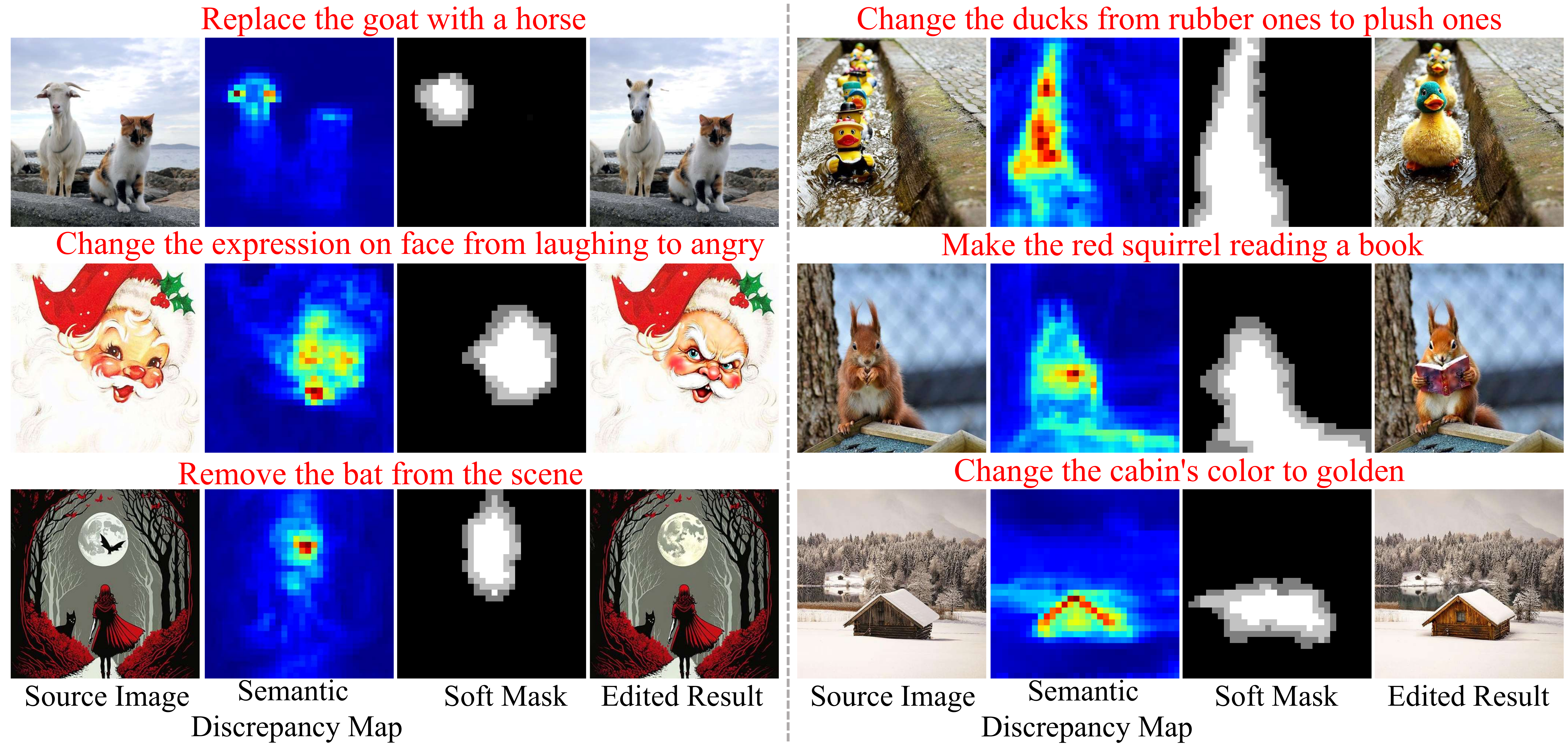}
    \caption{Visualization of semantic discrepancy map and soft mask.}
    \label{fig:more_visualization}
\end{figure}

\begin{figure}[t]
    \centering
    \includegraphics[width=\linewidth]{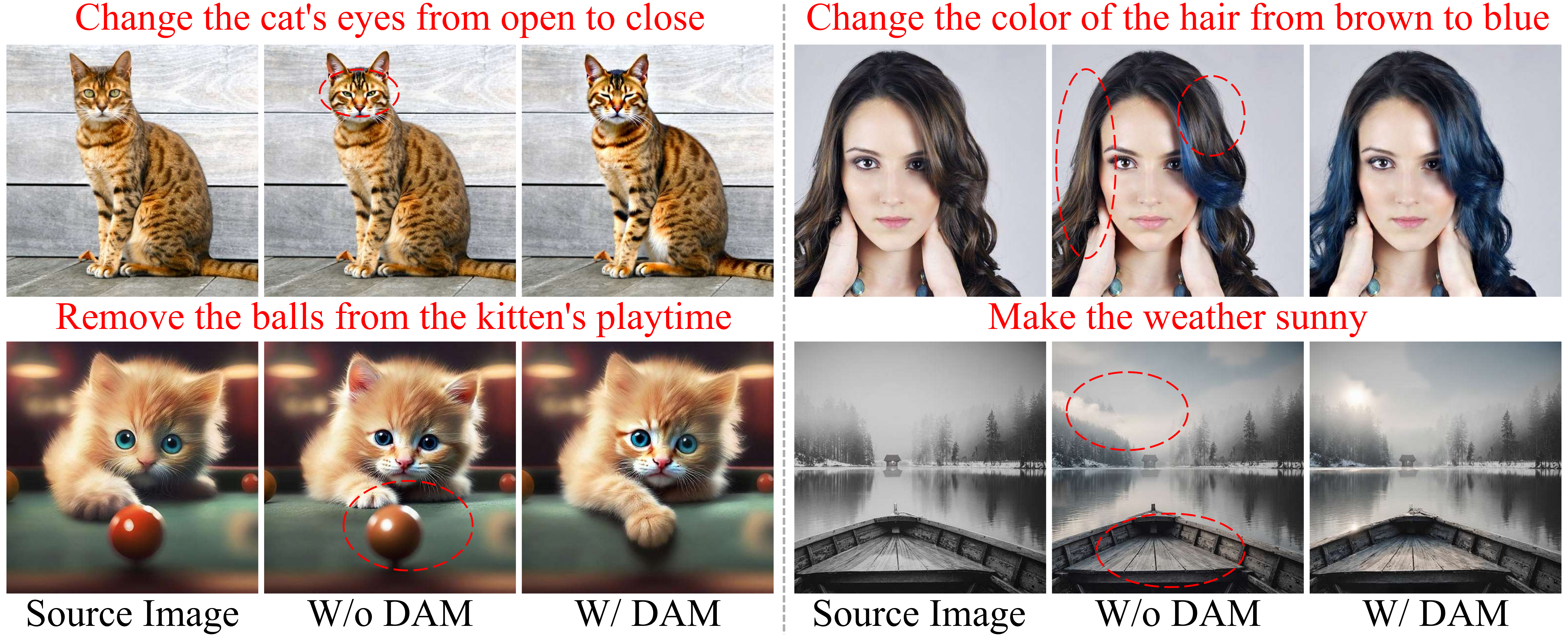}
    \caption{Effect of disparity-aware attention modulation (DAM).}
    \label{fig:ablation3}
\end{figure}

\subsection{Ablation Study}

\noindent \textbf{Effect of Adaptive Soft Mask.}
FusionEdit generates an adaptive soft mask to localize editing regions while ensuring smooth spatial transitions.
As shown in Table~\ref{tab:ablation1}, removing the mask results in a significant performance drop, particularly in background preservation metrics, highlighting the importance of spatial guidance.

We further compare our adaptive soft mask with a binary mask in Fig.~\ref{fig:ablation2}.
While binary masks can preserve background regions, they often produce artifacts near editing boundaries, such as residual structures from the source image that remain unmodified.
In contrast, our adaptive soft mask achieves better results by enabling gradual boundary transitions and precise localization, leading to coherent and artifact-free edits.

\noindent \textbf{Visualization of Semantic Discrepancy Map and Soft Mask.}
The semantic discrepancy map, derived from the difference between target- and source-conditioned velocity fields, guides the construction of the adaptive soft mask.
As shown in Fig.~\ref{fig:more_visualization}, the discrepancy map reliably highlights the regions requiring modification without relying on external cues.
After refinement, the resulting soft mask produces smooth boundary transitions, mitigating artifacts and leading to high-quality editing results.

\noindent \textbf{Effect of Disparity-Aware Attention Modulation (DAM).}
DAM transfers global appearance statistics from unmasked path into the editing path, enhancing editability and preserving global semantics.
As illustrated in Fig.~\ref{fig:ablation3}, removing DAM leads to restricted and inconsistent edits due to the suppression of cross-region cues by masking.
With DAM, editing becomes semantically accurate while preserving global coherence with the source, validating its effectiveness.

\section{Conclusion}
In this paper, we propose FusionEdit, a novel training-free framework for text-guided image editing with precision and controllability. 
By quantifying semantic discrepancies between source and target prompts, FusionEdit localizes editing and preserved regions. 
A distance-aware latent fusion with total variation regularization ensures smooth transitions across region boundaries.
Furthermore, FusionEdit designs AdaIN-based attention modulation to enhance editing consistency and preserve global semantics. 
Experiments demonstrate that FusionEdit outperforms state-of-the-art methods, achieving faithful preservation in preserved regions and precise alignment with the target prompt in editing regions.

\newpage

\footnotesize 
\section{Acknowledgments}
This work was supported by Guangdong Basic and Applied Basic Research Foundation under grants 2024A1515140109 and 2023A1515110695.

\bibliographystyle{IEEEbib}
\bibliography{strings,refs}

\end{document}